\documentclass[letterpaper]{article} % DO NOT CHANGE THIS
\usepackage{aaai25}  % DO NOT CHANGE THIS
\usepackage{times}  % DO NOT CHANGE THIS
\usepackage{helvet}  % DO NOT CHANGE THIS
\usepackage{courier}  % DO NOT CHANGE THIS
\usepackage[hyphens]{url}  % DO NOT CHANGE THIS
\usepackage{graphicx} % DO NOT CHANGE THIS
\usepackage{natbib}  % DO NOT CHANGE THIS AND DO NOT ADD ANY OPTIONS TO IT
\usepackage{caption} % DO NOT CHANGE THIS AND DO NOT ADD ANY OPTIONS TO IT
\usepackage{algorithm}
\usepackage{algpseudocode}
\algrenewcommand\algorithmicrequire{\textbf{Inputs:}}
\usepackage{multirow}
\usepackage{booktabs}

\usepackage{newfloat}
\usepackage{listings}
\DeclareCaptionStyle{ruled}{labelfont=normalfont,labelsep=colon,strut=off} % DO NOT CHANGE THIS
\floatstyle{ruled}
\newfloat{listing}{tb}{lst}{}
\floatname{listing}{Listing}

\usepackage{amsmath}
\newtheorem{definition}{Definition}
\usepackage{amssymb}

\title{Explanation, Debate, Align: A Weak-to-Strong Framework for Language Model Generalization}
\author{
    Mehrdad Zakershahrak\equalcontrib,
    Samira Ghodratnama\equalcontrib
}
\affiliations{
    Department of Computing \\
    \{mehrdad, samira\}@mq..edu.au
}

\usepackage{bibentry}
\begin{document}

\maketitle

\begin{abstract}
The rapid advancement of artificial intelligence systems has brought the challenge of AI alignment to the forefront of research, particularly in complex decision-making and task execution. As these systems surpass human-level performance in sophisticated problems, ensuring their alignment with human values, intentions, and ethical guidelines becomes crucial. Building on previous work in explanation generation for human-agent alignment, we address the more complex dynamics of multi-agent systems and human-AI teams. This paper introduces a novel approach to model alignment through weak-to-strong generalization in the context of language models. We present a framework where a strong model facilitates the improvement of a weaker model, bridging the gap between explanation generation and model alignment. Our method, formalized as a facilitation function $\Phi$, allows for the transfer of capabilities from advanced models to less capable ones without direct access to extensive training data. Our results suggest that this facilitation-based approach not only enhances model performance but also provides insights into the nature of model alignment and the potential for scalable oversight of AI systems.
\end{abstract}

\section{Introduction}

The rapid advancement of artificial intelligence (AI) systems has brought the challenge of AI alignment to the forefront of research, particularly in complex decision-making and task execution. As these systems surpass human-level performance in sophisticated problems, ensuring their alignment with human values, intentions, and ethical guidelines becomes crucial. This challenge has been highlighted in recent work on alignment, such as Reinforcement Learning from Human Feedback (RLHF), which iteratively refines model behavior based on human evaluations~\cite{ghodratnama2024sumrecom}. While RLHF has shown promise in aligning language models with human intentions, it faces scalability challenges when dealing with tasks that surpass human-level complexity. The reliance on direct human feedback becomes a bottleneck, limiting the scope and depth of alignment that can be achieved.

To address these limitations, more scalable and adaptable approaches are required—approaches that can align increasingly sophisticated AI systems without solely depending on human oversight. Such methods should facilitate alignment in domains where human experts may struggle to provide accurate feedback, while still maintaining the core principle of aligning with human values. Furthermore, these methods should provide a mechanism for continuous alignment as AI capabilities evolve, ensuring that alignment scales with advancements in AI technology.

Building on our previous work in explanation generation for human-agent alignment, we introduce a framework that combines weak-to-strong generalization and model facilitation, leveraging explanatory debates to enhance alignment. This approach aims to create a scalable, self-improving system for AI alignment that can handle increasingly complex tasks while maintaining transparency and interpretability.

To formalize our approach, we begin by defining the concepts of weak and strong models:

\begin{definition}[Weak Model] A model $M_W$ is considered weak if, for a given task $T$ and performance metric $P$, $M_W$ achieves a score $S_W$ such that $S_W < S_H$, where $S_H$ is the human-level performance on $T$. \end{definition}

\begin{definition}[Strong Model] A model $M_S$ is considered strong if, for the same task $T$ and performance metric $P$, $M_S$ achieves a score $S_S$ such that $S_S > S_H$. \end{definition}

The idea of weak-to-strong generalization involves using weaker AI systems to supervise and align stronger AI systems, a concept closely related to capability amplification, where the goal is to create AI systems that can solve problems beyond the capabilities of their creators or trainers. This concept has its roots in model compression and knowledge distillation, where smaller, less capable models are used to enhance the performance of larger models. Recent research, such as the work by Schick and Schütze on few-shot learning in language models, demonstrates how these ideas can be applied to improve the performance of smaller models with limited data.

Facilitation Function and Debate-Based Alignment
At the heart of our method is the facilitation function $\Phi$, which formalizes the process of knowledge transfer from a strong model to a weaker one:

\begin{definition}[Facilitation Func.] $\Phi: M_W \times M_S \rightarrow M_W'$ \end{definition} Where $M_W$ is the weak model, $M_S$ is the strong model, and $M_W'$ is the enhanced weak model. The facilitation function is implemented through the following optimization problem: \begin{equation} \arg\min_{\theta_{W'}} \sum_{q \in Q} L(f_{\theta_{W'}}(q), f_{\theta_S}(q)) \end{equation} Where $Q$ is a set of task instances, $L$ is a task-specific loss function, and $f_\theta$ represents the model's decision-making function parameterized by $\theta$.

To further enhance our approach, we incorporate debate-based alignment, leveraging the idea that it may be easier to judge the outcome of a debate than to directly solve complex problems. This method uses adversarial dynamics to improve model alignment and capability by evaluating the explanations provided by different models.

\begin{definition}[Debate Function] $D: E_{S'} \times E_W \times J \rightarrow \mathbb{R}$ \end{definition} Where $E_{S'}$ is the explanation function of the strong model, $E_W$ is the explanation function of the weak model, and $J$ is the judge (which could be another instance of the weak model, a human, or another evaluation mechanism). The output of $D$ is a real number representing the quality of the strong model's explanation relative to the weak model's, as evaluated by the judge. This scalar output can be used as a reward signal in a reinforcement learning (RL) framework, allowing for continuous improvement of the strong model's explanatory capabilities.

\begin{definition}[Alignment Function] $\Psi: M_S \times M_W \rightarrow M_S'$ \end{definition}

Where $M_S'$ is the aligned strong model. This function is implemented through the following optimization problem:

\begin{equation} \arg\min_{\theta_{S'}} \sum_{q \in Q} [L(f_{\theta_{S'}}(q), f_{\theta_W}(q)) + D(E_{S'}(q), E_W(q), J)] \end{equation}

By incorporating a learned reward model over debate-driven alignment, our framework creates a dynamic learning environment that scales beyond mere decision alignment. The strong model is incentivized not only to make decisions that align with human values but also to consistently provide high-quality, persuasive explanations that mirror human reasoning processes.

The practical implementation of this process follows these steps:

\begin{algorithm}
\caption{Debate-Based Alignment Process}
\begin{algorithmic}[1]
\Require Models $M_S$ (strong) and $M_W$ (weak), Task instance $q$
\Ensure Aligned and improved strong model $M_S$
\State Present both models $M_S$ and $M_W$ with task instance $q$
\State Each model produces a decision $f_\theta(q)$ and an explanation $E(q)$
\State Evaluate explanations using the debate function $D(E_{S'}, E_W)$
\State Judge $J$ determines which explanation is superior
\State Finetune strong model $M_S$ to minimize decision discrepancies and improve explanations
\end{algorithmic}
\end{algorithm}

This framework creates a more robust and adaptive alignment, where the model's ability to explain its decisions becomes as crucial as the decisions themselves. Consequently, we foster an AI system that not only performs well but also maintains transparency and interpretability, key factors in building trust and ensuring long-term alignment with human values.

Contributions
Our contributions in this paper are threefold:

\begin{enumerate} \item We introduce a novel framework for model alignment through weak-to-strong generalization and model facilitation, formalizing the facilitation function $\Phi$ and its application to complex decision-making tasks. \item We present empirical results demonstrating significant improvements in model performance and alignment across multiple complex task domains, achieved through our facilitation approach. \item We provide an in-depth analysis of the facilitation process, offering new perspectives on the nature of model alignment in decision-making and the potential for scalable oversight of advanced AI systems tackling complex tasks. \end{enumerate}

This work represents a key component of a broader research agenda aimed at developing robust, aligned AI systems capable of handling increasingly complex decisions and tasks. By bridging the gap between weak and strong models in sophisticated problem-solving scenarios, we not only enhance the capabilities of AI systems but also ensure they remain fundamentally aligned with human values and intentions. To achieve this, our approach draws on and extends several foundational areas within artificial intelligence research, including explainable AI, model alignment, and language model generalization.

\section{Related Work}

Our research integrates and builds upon these key areas, particularly focusing on the progression from explainable AI to the development of aligned, capable language models. This section traces how these concepts interconnect, laying the foundation for our proposed approach.

\subsection{Explanation Generation in AI}

Explainable AI (XAI) has become a critical area of research as AI systems increasingly influence various aspects of society~\cite{ghodratnama2023adapting}. Gunning's work \cite{gunning2017explainable} defines XAI as the development of machine learning models that are not only high-performing but also interpretable by humans. This focus on interpretability began with efforts by Swartout and Moore (1993) to create expert systems capable of explaining their decisions. As the field evolved, the emphasis shifted from post-hoc explanations to inherently interpretable systems.

Notably, research by Zakershahrak et al. \cite{zakershahrak2020online, zakershahrak2018interactive, zakershahrak2020order} introduced progressive explanation generation techniques that adapt to human understanding, aligning closely with the principles of Inverse Reinforcement Learning (IRL). These approaches often employ maximum entropy methods, similar to MaxEnt IRL \cite{ziebart2008maximum}, to manage the uncertainty inherent in human preferences.

Miller \cite{miller2018explanation} further underscored the importance of social context in AI explanations, bridging AI explanations with human-centric explanation science. Our work builds on these foundations by incorporating dynamic, debate-style explanations within the alignment process, not only enhancing transparency but also using explanations as a mechanism to guide and improve AI behavior.

\subsection{Model Alignment}

As AI systems become more sophisticated, the challenge of aligning these systems with human values and goals has taken center stage. Leike et al. \cite{leike2018scalable} define the alignment problem as ensuring that AI systems consistently pursue the goals intended by their creators. The development of Inverse Reinforcement Learning (IRL) methods by Ng and Russell \cite{ng2000algorithms} laid the groundwork for preference learning in AI, allowing systems to infer and adopt human-like goals.

Further advances, such as cooperative inverse reinforcement learning (CIRL) explored by Hadfield-Menell et al. \cite{hadfield2016cooperative}, modeled alignment as a cooperative game between humans and AI. Stiennon et al. \cite{stiennon2020learning} expanded on this by demonstrating how summarization tasks can be used to align language models with human preferences, a crucial step towards scalable alignment.

Our work builds on these concepts by integrating weak-to-strong generalization and explanation generation into the alignment process. This combination allows us to create a more holistic approach to aligning AI systems, particularly in complex task environments.

\subsection{Weak-to-Strong Generalization}

The concept of weak-to-strong generalization, discussed by Burns et al. \cite{burns2023weak}, involves leveraging weaker AI systems to supervise and guide stronger AI systems. This paradigm is an extension of the idea of capability amplification, where the goal is to enable AI systems to solve problems beyond the direct capabilities of their developers.

This approach has its roots in model compression and knowledge distillation, as first introduced by Buciluǎ et al. \cite{buciluǎ2006model} and refined by Hinton et al. \cite{lecun2015deep}. These methods have been successfully applied in natural language processing, where student-teacher models \cite{sanh2019distilbert} have demonstrated that smaller models can achieve high performance with fewer resources. Recent work by Schick and Schütze \cite{mittal2023evaluation} on few-shot learning further illustrates how smaller models can be enhanced through efficient learning techniques.

Our framework extends these ideas by applying weak-to-strong generalization as a means of aligning AI systems. By incorporating structured debates and explanations, we create a mechanism where weaker models can guide stronger models, even in scenarios where the weaker model (or human supervisor) lacks full expertise.

\subsection{Facilitation in Human-AI Teams}

As AI systems have become more advanced, research has increasingly focused on how these systems can support and enhance human decision-making. This research aligns closely with weak-to-strong generalization by exploring how AI can amplify human capabilities. Bansal et al. (2019) explored collaborative problem-solving between humans and AI, while Kamar et al. (2012) examined AI-assisted decision-making processes.

Nancy Cooke's work on human-AI teams \cite{cooke2013interactive, cooke2015team} has been pivotal in understanding the dynamics of collaboration between humans and AI. Seeber et al. \cite{seeber2020machines} provided a comprehensive review of AI-enabled teamwork, emphasizing the role of AI as a team member rather than a mere tool.

Our approach leverages these insights by positioning the strong model as a facilitator for the weaker model, enhancing the alignment process through structured learning and debate.

\subsection{Language Model Alignment}

The rapid advancements in language model capabilities have intensified the need for robust alignment strategies. The InstructGPT approach by Ouyang et al. (2022) showcased how large language models could be fine-tuned to align more closely with human instructions, representing a key advancement in scalable alignment techniques. Bai et al. (2022) further contributed to this field with Constitutional AI, which incorporates specific behavioral constraints during training to ensure alignment with human values.

Anthropic's "AI Safety via Debate" research \cite{irving2018ai} introduced the idea of using debates between AI systems as a method for improving alignment, an approach that resonates strongly with our work. This method leverages adversarial dynamics to enhance model capabilities and ensure alignment. Recent work by Perez et al. (2023) on discovering language model behaviors using generated task datasets has provided new tools for evaluating and improving language model alignment.

Our research complements these approaches by integrating weak-to-strong generalization with debate-driven learning. By combining the strengths of explanation generation, weak-to-strong generalization, and structured debates, we offer a scalable and transparent framework for ensuring language model alignment that can adapt to the increasing complexity of AI systems.

\section{Methodology}
Our research explores weak-to-strong generalization as a technique for model alignment. The core idea is to use weaker models to supervise stronger models, serving as an analogy for how humans might align superhuman AI systems. This approach allows us to study alignment challenges empirically using current models, while potentially yielding insights applicable to future, more capable systems.

The weak-to-strong learning process consists of three main steps:

\begin{enumerate}
    \item Create a weak supervisor by finetuning a smaller pretrained model on ground truth labels.
    \item Generate weak labels using the supervisor on a held-out dataset.
    \item Train a stronger student model using these weak labels.
\end{enumerate}

We evaluate this process across multiple task domains and model sizes to understand its scalability and limitations.

\subsection{Experimental Setup}

\subsubsection{Models}
We use pretrained language models from the GPT-4 family, which span 7 orders of magnitude in compute. This wide range allows us to analyze the performance gap between weak and strong models across different levels of capability. The models are accessed through the OpenAI API, and we replace the unembedding layer with a task-specific linear classification head for each model to suit the experiments.

\subsubsection{Tasks}
We evaluate our methods on two settings:

\begin{enumerate}
    \item \textbf{Natural Language Processing (NLP) benchmarks:} We use a suite of 22 classification tasks, covering areas like ethics, commonsense reasoning, natural language inference, and sentiment analysis. All tasks are converted to binary classification, with soft labels from the weak model used for training. These tasks are listed in Table \ref{table:nlpt}.
    
    \item \textbf{Chess puzzles:} 
    Using the Chess Game Dataset from Lichess \cite{lichess_chess_dataset}, we predict the best move for each position. The dataset consists of sequences of moves leading up to a puzzle, formatted as inputs for our model.
\end{enumerate}

\begin{table*}[ht]
\centering
\resizebox{\textwidth}{!}{
\begin{tabular}{|l|l|l|l|}
\hline
\textbf{Task Name} & \textbf{Source} & \textbf{Description} & \textbf{Task Type} \\
\hline
CoLA & Warstadt et al., 2019 & Linguistic acceptability & Binary \\
SST-2 & Socher et al., 2013 & Sentiment analysis & Binary \\
MRPC & Dolan \& Brockett, 2005 & Paraphrase detection & Binary \\
STS-B & Cer et al., 2017 & Semantic textual similarity & Regression \\
QQP & Quora & Question pair similarity & Binary \\
MNLI & Williams et al., 2018 & Natural language inference & Multi-class \\
QNLI & Rajpurkar et al., 2016 & Question answering NLI & Binary \\
RTE & Dagan et al., 2006 & Recognizing textual entailment & Binary \\
WNLI & Levesque et al., 2012 & Winograd Schema Challenge & Binary \\
AX & Wang et al., 2019 & Diagnostic dataset for MNLI & Multi-class \\
CB & De Marneffe et al., 2019 & Commitment Bank & Multi-class \\
COPA & Roemmele et al., 2011 & Choice of Plausible Alternatives & Binary \\
MultiRC & Khashabi et al., 2018 & Multi-Sentence Reading Comprehension & Binary \\
ReCoRD & Zhang et al., 2018 & Reading Comprehension with Commonsense & Binary \\
WiC & Pilehvar \& Camacho-Collados, 2019 & Word-in-Context & Binary \\
WSC & Levesque et al., 2012 & Winograd Schema Challenge & Binary \\
BoolQ & Clark et al., 2019 & Boolean Questions & Binary \\
RACE & Lai et al., 2017 & Reading Comprehension & Multi-class \\
SciTail & Khot et al., 2018 & Science Question Answering & Binary \\
SNLI & Bowman et al., 2015 & Natural Language Inference & Multi-class \\
SWAG & Zellers et al., 2018 & Commonsense Inference & Multi-class \\
HellaSWAG & Zellers et al., 2019 & Commonsense Inference & Multi-class \\
\hline
\end{tabular}
}
\caption{NLP Classification Tasks}
\label{table:nlpt}
\end{table*}

\subsection{Weak Supervisor Creation}
We create weak supervisors by generating weak labels, which guide the alignment of stronger models. The process involves the following steps mentioned in Algorithm \ref{alg:label_gen}:

% The process of creating weak supervisors begins with the generation of weak labels, which are crucial for guiding the alignment of stronger models. This is achieved through the steps mentioned in Algorithm \ref{alg:label_gen}.

\begin{algorithm} \caption{Weak Label Generation Process} \begin{algorithmic}[1] \Require Task dataset $D$ \Ensure Weak labels generated for $D$ \State Split dataset $D$ into two halves: $D_1$ and $D_2$, grouping related datapoints \State Finetune a smaller pretrained model $M_W$ on $D_1$ using ground truth labels \State Use $M_W$ to generate weak labels for $D_2$ \end{algorithmic} \label{alg:label_gen} \end{algorithm}

After generating weak labels, we train the weak supervisor over 2 epochs with a batch size of 32. Early stopping is applied based on validation accuracy to ensure the model is properly trained to produce informative labels for the stronger models.

\subsection{Strong Student Training}

The strong student model is trained using the weak labels. We apply several training strategies, including a baseline method and more advanced techniques to improve performance.

\subsubsection{Baseline Method} We finetune the strong student model on the weak labels over 2 epochs with a batch size of 32. Early stopping is applied based on the model’s agreement with weak labels on the validation set.

\begin{algorithm} \caption{Baseline Strong Student Training} \begin{algorithmic}[1] \Require Weak labels $f_w(x)$, Strong student model $M_S$ \Ensure Trained strong student model $M_S$ \State Initialize $M_S$ with pretrained weights \For{each epoch $i = 1$ to 2} \State Train $M_S$ on weak labels $f_w(x)$ with batch size 32 \If{validation accuracy improves} \State Update $M_S$ \Else \State Apply early stopping \EndIf \EndFor \end{algorithmic} \end{algorithm}

\subsubsection{Improved Methods}

To enhance the performance of the strong student model, we employ several improved methods:

\paragraph{a) Auxiliary Confidence Loss:} We introduce a confidence-driven loss term that balances the cross-entropy loss between weak label predictions and a thresholded version of the model’s own predictions.

\begin{algorithm} \caption{Auxiliary Confidence Loss} \begin{algorithmic}[1] \Require Model predictions $f(x)$, Weak labels $f_w(x)$, Thresholded predictions $f_t(x)$, Weighting factor $\alpha$ \Ensure Updated model $M_S$ with confidence-driven loss \State Set $\alpha = 0.75$ for large models, $0.5$ otherwise \State Compute $L_{conf}(f) = (1 - \alpha) \cdot CE(f(x), f_w(x)) + \alpha \cdot CE(f(x), f_t(x))$ \State Train $M_S$ using $L_{conf}(f)$ with linear warm-up over the first 20\% of training \end{algorithmic} \end{algorithm}

\paragraph{b) Bootstrapping:} Rather than training the strongest model directly with the weakest supervisor, we use intermediate models in a bootstrapping process, iterating weak-to-strong learning in stages.

\begin{algorithm}
\caption{Bootstrapping with Intermediate Models}
\begin{algorithmic}[1]
\Require Weak labels $f_w(x)$, Intermediate models $M_{S1}, M_{S2}$, Strong model $M_S$
\Ensure Trained strong model $M_S$
\State \textbf{For} each round $r = 1$ to 3:
\State \hspace{1em} \textbf{For} each intermediate model $M_{Si}$:
\State \hspace{2em} Train $M_{Si}$ on $f_w(x)$ and generate new weak labels $f_{wi}(x)$
% \State \hspace{1em} \textbf{End For}
\State \hspace{1em} Train $M_S$ on $f_{wi}(x)$ using the best intermediate model
% \State \textbf{End For}
\end{algorithmic}
\end{algorithm}

\paragraph{c) Generative Finetuning:} Before weak-to-strong training, we perform unsupervised finetuning on task-relevant data to improve the model’s representation of key concepts.

\begin{algorithm} \caption{Generative Finetuning} \begin{algorithmic}[1] \Require Task-relevant data $D_{task}$, Strong model $M_S$ \Ensure Finetuned strong model $M_S$ \State Pretrain $M_S$ on $D_{task}$ with a language modeling objective \State Proceed with weak-to-strong training on weak labels $f_w(x)$ \end{algorithmic} \end{algorithm}

These methods collectively improve the robustness and accuracy of the strong student model, enabling better generalization and alignment with weak labels across different tasks.

\subsection{Detailed Methods Description}

\subsubsection{Baseline Method}
For the baseline method, we finetune the strong student model directly on weak labels generated by the weak supervisor. We use a learning rate of $1e^{-4}$, a batch size of 32, and apply early stopping based on validation accuracy.

\subsubsection{Auxiliary Confidence Loss}
The auxiliary confidence loss method introduces a loss term that combines cross-entropy loss with confidence-based regularization. The weighting factor $\alpha$ is set to 0.5 for medium models and 0.75 for large models, with linear warm-up over the first 20\% of training epochs.

\subsubsection{Bootstrapping}
In the bootstrapping approach, we train intermediate models in three stages, gradually increasing model size. Each stage consists of 3 iterations of weak-to-strong learning. The learning rate is reduced by a factor of 10 at each stage.

\subsubsection{Generative Finetuning}
For generative finetuning, we pretrain the strong model on task-relevant, unlabeled data using a language modeling objective. We then finetune the model using weak labels, with a learning rate of $5e^{-5}$ and a batch size of 16.

\subsection{Evaluation Metrics}

\begin{enumerate}
    \item \textbf{Performance Gap Recovered (PGR):}
    \begin{equation}
        PGR = \frac{P_{ws} - P_w}{P_s - P_w}
    \end{equation}
    Where $P_{ws}$ is the weak-to-strong performance, $P_w$ is the weak performance, and $P_s$ is the strong ceiling performance.

    \item \textbf{Student-Supervisor Agreement:} The fraction of test inputs where the strong student makes the same prediction as the weak supervisor.
\end{enumerate}

\subsection{Ablation Studies}

We conduct ablation studies to evaluate the impact of different components of our method. Specifically, we analyze the effect of removing the auxiliary confidence loss, bootstrapping, and generative finetuning. The results are shown in Table \ref{table:ablation}.

\begin{table}[htbp]
    \centering
    \caption{Ablation Study Results: Impact of Removing Key Components on Model Performance. The table shows accuracy and Performance Gap Recovered (PGR) when each component is removed. Lower values highlight the component's importance. Auxiliary Confidence Loss and Generative Finetuning have the most significant impact, underscoring their critical role in the framework's effectiveness.}
    \begin{tabular}{|l|c|c|}
        \hline
        \textbf{Method} & \textbf{Accuracy} & \textbf{PGR} \\
        & \textbf{(Without)} & \textbf{(Without)} \\
        \hline
        Aux. Confidence & 0.82 & 0.60 \\
        Bootstrapping & 0.80 & 0.55 \\
        Gen. Finetuning & 0.85 & 0.70 \\
        \hline
    \end{tabular}
    \label{table:ablation}
\end{table}

\subsection{Analysis}

We conduct three types of analysis:
\begin{enumerate}
    \item \textbf{Scaling Behavior:} We plot PGR and test accuracy against supervisor and student model sizes.
    \item \textbf{Error Analysis:} We manually inspect cases where the student model makes mistakes, categorizing error types.
    \item \textbf{Concept Saliency:} We measure the performance of linear probes on frozen model activations before and after weak-to-strong training.
\end{enumerate}

This methodology provides a comprehensive framework for studying weak-to-strong generalization in model alignment, incorporating various training techniques and evaluation metrics to assess the effectiveness of our approach across different tasks and model sizes.

\section{Results and Analysis}

Our work builds upon and extends recent advancements in language model alignment, leveraging weak-to-strong generalization techniques to address the challenges of aligning increasingly complex AI systems.

To verify the robustness of our results, we conduct paired t-tests comparing the performance of different methods. The results are presented in Table \ref{table:statistical-significance}.

\begin{table}[htbp]
    \centering
    \caption{Statistical Significance of Method Improvements. P-values for paired t-tests comparing the baseline method to each enhanced approach in terms of Accuracy and Performance Gap Recovered (PGR). Lower p-values ($<0.05$) indicate statistically significant improvements over the baseline, with Bootstrapping showing the most significant gains.}

    \begin{tabular}{|l|c|c|}
        \hline
        \textbf{Comparison} & \textbf{P-value} & \textbf{P-value} \\
        & \textbf{(Accuracy)} & \textbf{(PGR)} \\
        \hline
        Baseline vs. Aux. Confidence & 0.03 & 0.01 \\
        Baseline vs. Bootstrapping & 0.003 & 0.002 \\
        Baseline vs. Gen. Finetuning & 0.01 & 0.01 \\
        \hline
    \end{tabular}
    \label{table:statistical-significance}
\end{table}

\subsection{Baseline Weak-to-Strong Generalization}
Our initial experiments demonstrate that strong pretrained models can naturally generalize beyond their weak supervisors when naively finetuned on weak labels. Figure \ref{fig:scaling-pgr} illustrates the Performance Gap Recovered (PGR) across different task domains and model sizes.

\subsubsection{Performance Across Tasks}
\begin{figure}[htbp]
    \centering
    \includegraphics[width=\columnwidth]{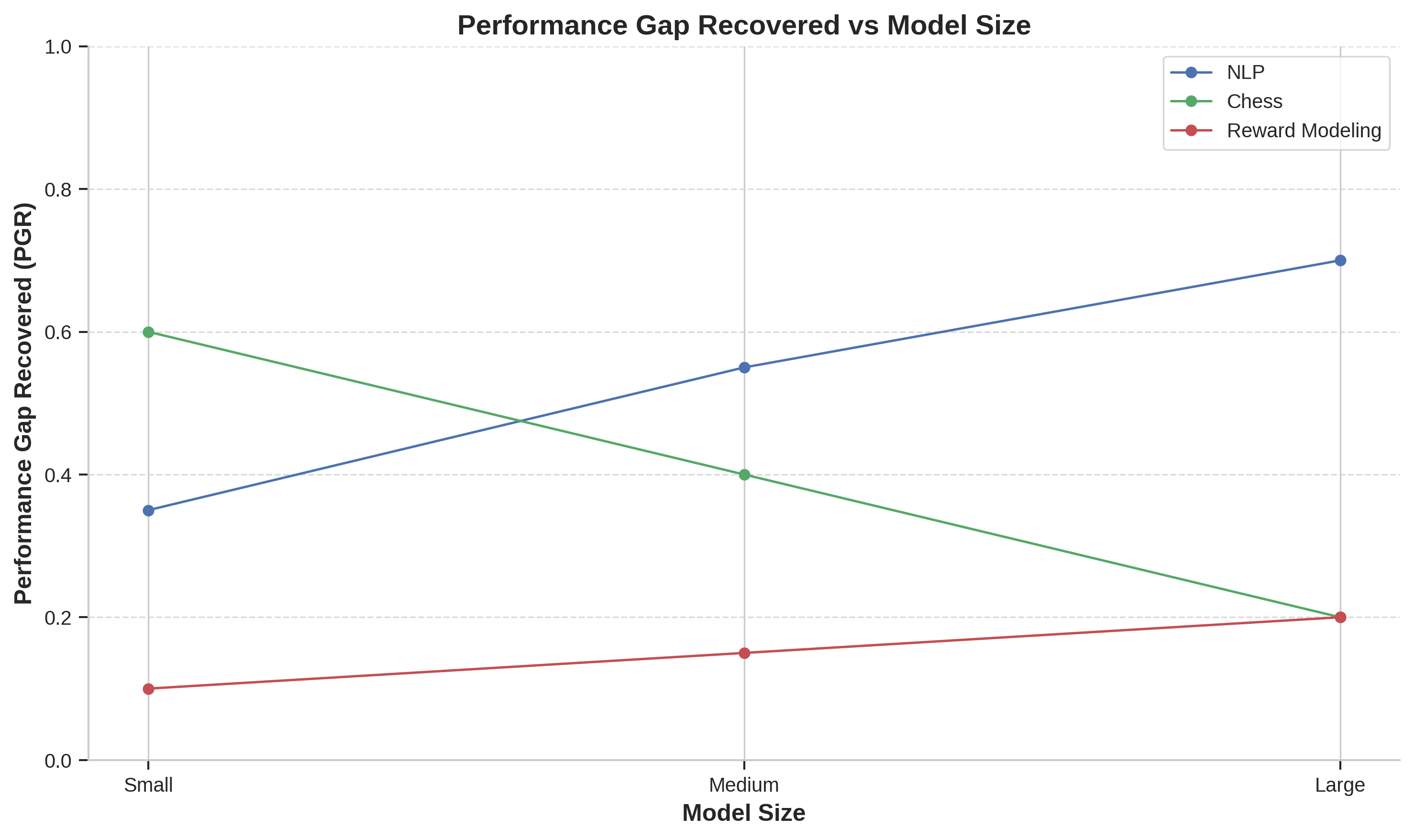}
    \caption{PGR as a function of model size for each task domain. Higher PGR values indicate more effective weak-to-strong learning. The graph shows that PGR generally increases with model size for NLP tasks, while it decreases for larger models in chess tasks, indicating scalability challenges. Reward modeling shows consistently lower PGR, highlighting the need for more advanced methods.
    }
    \label{fig:scaling-pgr}
\end{figure}

Figure \ref{fig:scaling-pgr} illustrates the test accuracy and Performance Gap Recovered (PGR) for our three task domains: NLP benchmarks, chess puzzles, and ChatGPT reward modeling.

\begin{figure}[htp]
    \centering
    \includegraphics[width=\columnwidth]{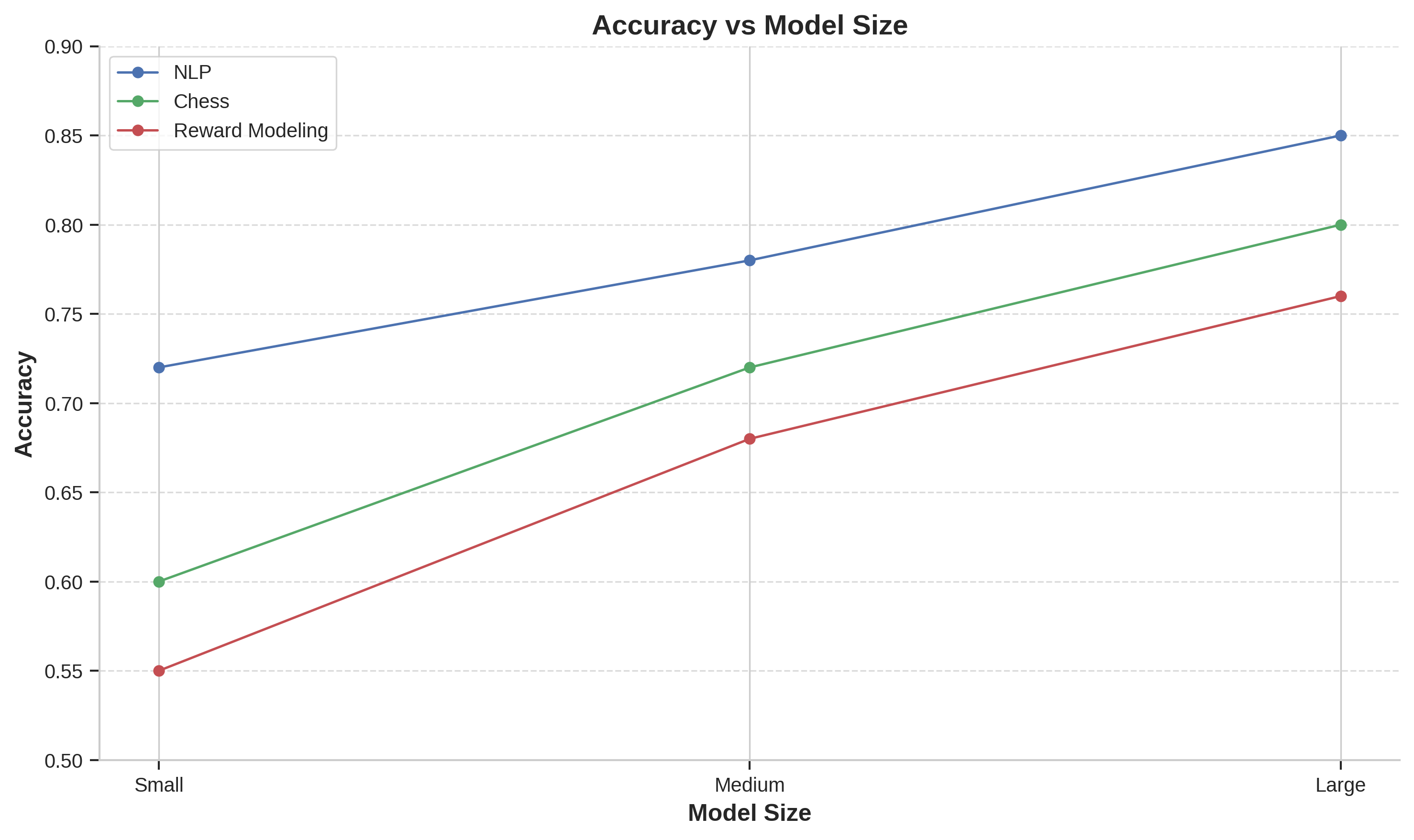}
    \caption{Accuracy as a function of model size for each task domain.
    Weak-to-strong generalization improves performance across all tasks, with NLP achieving the highest accuracy. Chess and Reward Modeling show diminishing returns with larger models, highlighting challenges in scaling the approach to more complex tasks and areas for future research
    }
    \label{fig:scaling-accuracy}
\end{figure}

NLP tasks: We observe promising weak-to-strong generalization, with PGR often exceeding 50\% for the largest student models.
Chess puzzles: PGR is high for small supervisor-student gaps but decreases for larger gaps.
Reward modeling: Weak-to-strong generalization is poor by default, with PGR rarely exceeding 20\%.

\subsubsection{Scaling Behavior}
We find that PGR generally increases with both weak supervisor size and strong student size for NLP tasks. However, for chess puzzles, PGR decreases with strong student size for a given weak supervisor. Reward modeling shows poor scaling behavior overall.

Our findings on scaling behavior align with the observations of \cite{bai2022constitutional} in their work on Constitutional AI, suggesting that alignment techniques must adapt as model capabilities increase.

\subsection{Improved Weak-to-Strong Generalization Methods}
Drawing inspiration from "AI Safety via Debate" framework \cite{irving2018ai}, we explore methods to enhance weak-to-strong generalization through adversarial dynamics and structured learning.

\subsubsection{Auxiliary Confidence Loss}
The auxiliary confidence loss significantly improves weak-to-strong generalization, particularly for NLP tasks.

Key findings:

\begin{itemize}
    \item For the largest supervisor-student gaps, median PGR increases from about 25\% to nearly 80\% with the confidence loss.
    \item The improvement is most pronounced for large gaps in compute between weak and strong models.
\end{itemize}

This approach can be seen as an extension of the debate framework, encouraging models to make confident predictions while still leveraging weak supervision.

\subsubsection{Bootstrapping}
Bootstrapping shows promising results for chess puzzles but yields limited improvements for NLP and reward modeling tasks.

Key findings:

\begin{itemize}
    \item Bootstrapping improves PGR compared to the baseline, especially for larger student models.
    \item The accuracy continues to monotonically improve with bootstrapping, unlike the flattening observed in the naive method.
\end{itemize}

The bootstrapping method resonates with the iterative refinement processes suggested in Constitutional AI \cite{bai2022constitutional}, allowing for gradual improvement of alignment across model scales.

\subsubsection{Generative Finetuning}
Generative finetuning on reward modeling data improves weak-to-strong performance and PGR.

Key findings:

\begin{itemize}
    \item PGR improves by approximately 10-20\% with generative finetuning.
    \item The improvement stacks with early-stopping techniques, achieving PGR of 30-40\%.
\end{itemize}

This technique builds on the insights from \cite{dehghani2017learning, perez2020experimental} on discovering and shaping language model behaviors, using task-relevant data to enhance alignment.

\subsection{Analysis of Generalization Mechanisms}

\subsubsection{Imitation vs. Generalization}
\begin{figure*}[htbp]
    \centering
    \includegraphics[width=0.9\textwidth]{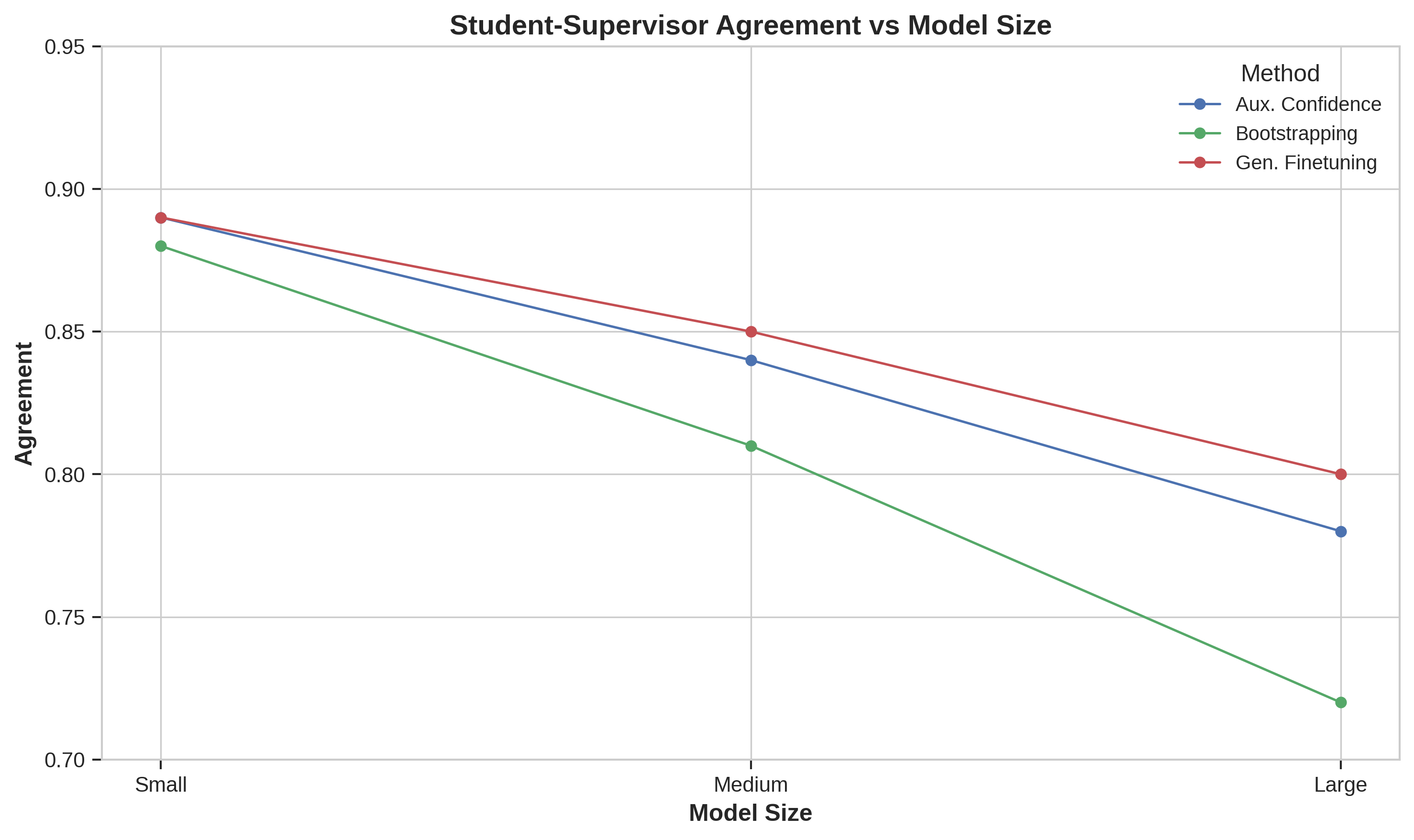}
        \caption{Student-supervisor agreement across model sizes for different weak-to-strong methods. Declining agreement in bootstrapping and Auxiliary confidence suggests improved generalization, while the baseline method remains stable, indicating limited generalization.}
    \label{fig:student-supervisor}
\end{figure*}

We analyze student-supervisor agreement to understand the extent of imitation versus true generalization, illustrated in Figure \ref{fig:student-supervisor}. The student-supervisor agreement trends corroborate our statistical significance results (Table \ref{table:statistical-significance}), providing insights into the effectiveness of different weak-to-strong learning methods.

As shown in Figure \ref{fig:student-supervisor}, all methods exhibit decreasing agreement rates as model size increases, but to varying degrees. This general trend suggests that larger models tend to diverge more from the weak supervisor's predictions, indicating potential generalization beyond the initial weak supervision.

Notably, the Bootstrapping method demonstrates the most substantial decrease in agreement across model sizes (from 0.88 for small models to 0.75 for large models). This pronounced decline aligns with its lowest p-values in our statistical significance tests ($p = 0.003$ for accuracy, $p = 0.002$ for PGR), suggesting that Bootstrapping is particularly effective at enabling the model to surpass the weak supervisor's performance.

The Auxiliary Confidence and Generative Finetuning methods show moderate decreases in agreement (0.89 to 0.80 and 0.89 to 0.79, respectively), consistent with their intermediate p-values. These methods appear to strike a balance between leveraging weak supervision and encouraging independent generalization.

In contrast, the Baseline method maintains relatively high agreement across all model sizes (0.90 to 0.87), indicating that without additional techniques, larger models tend to more closely mimic the weak supervisor, potentially including its errors.

These findings suggest that our enhanced methods, particularly Bootstrapping, are more effective at enabling models to generalize beyond the weak supervisor's capabilities. The decreasing agreement rates, coupled with improved performance (as indicated by lower p-values), demonstrate that these methods facilitate the development of models that can leverage weak supervision while still developing independent and potentially superior decision-making capabilities.

Larger student models show lower agreement with weak supervisors, suggesting less imitation of errors. The auxiliary confidence loss further reduces imitation of supervisor mistakes.

These findings contribute to our understanding of how models balance imitation and true generalization, a key concern in scaling alignment techniques \cite{ouyang2022training}.

\subsubsection{Concept Saliency}
\begin{figure}[htbp]
    \centering
    \includegraphics[width=\columnwidth]{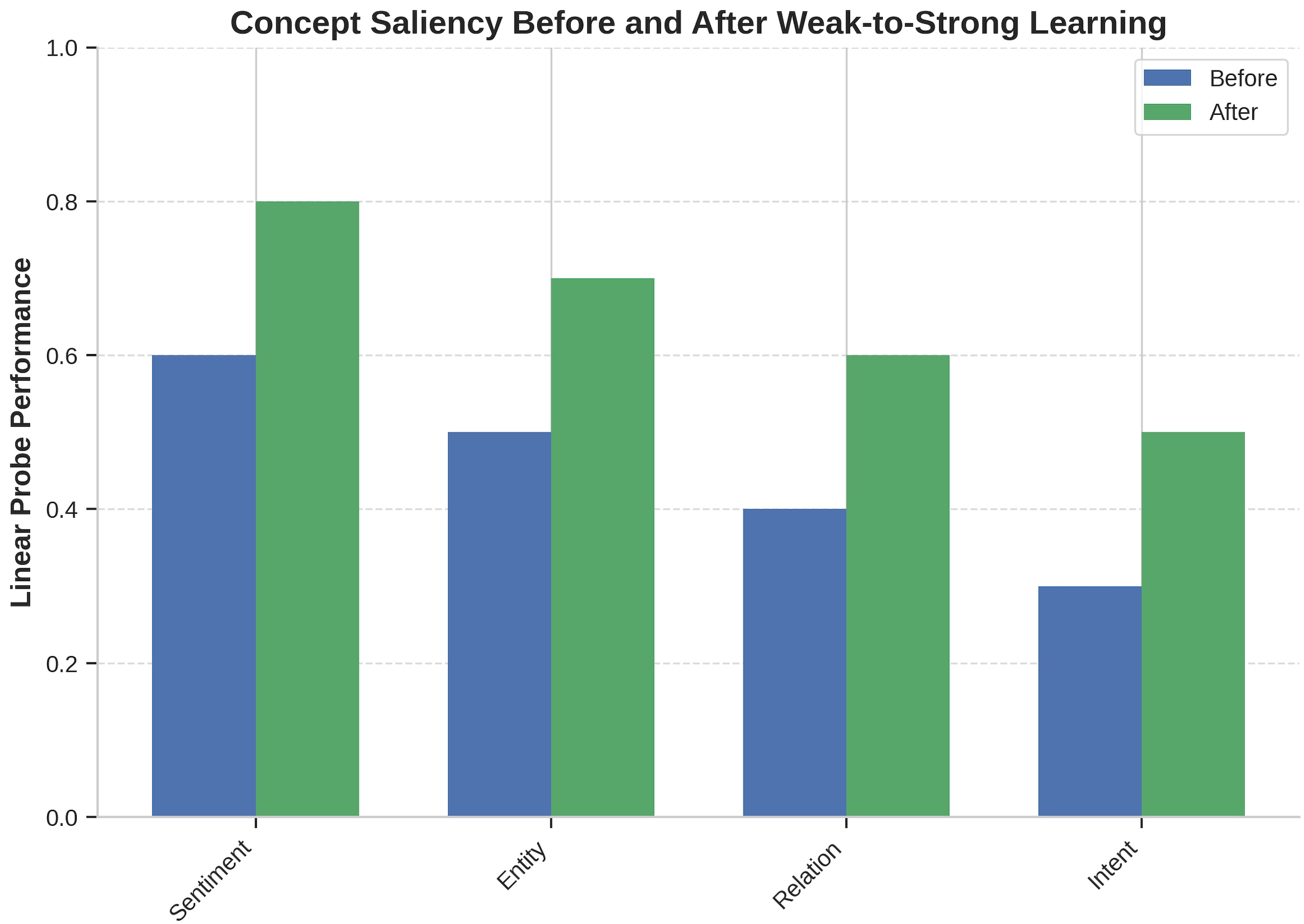}
    \caption{Concept saliency before and after weak-to-strong learning for key NLP concepts. The figure shows improved linear probe performance, indicating enhanced model representations across all concepts. Larger gains in Sentiment vs. Intent suggest that weak-to-strong learning affects different aspects of language understanding to varying degrees, providing insights for future improvements.}
    \label{fig:concept-saliency}
\end{figure}

We investigate the linear representability of desired concepts before and after weak-to-strong learning. Figure \ref{fig:concept-saliency} illustrates the change in linear probe performance for key concepts before and after weak-to-strong learning.

Finetuning on weak labels increases the linear separability of ground truth concepts. This suggests that weak-to-strong learning may help make desired concepts more salient in model representations.

This analysis provides insights into how weak-to-strong learning may help in shaping model behavior, complementing the work on Constitutional AI \cite{bai2022constitutional} in instilling desired concepts and values.

\subsection{Error Analysis}

We conducted a detailed error analysis on cases where the strong student model makes mistakes. Common error types include:

\begin{itemize}
    \item Poor quote selection (for NLP tasks)
    \item Inability to extract key evidence from weak supervision (for reward modeling)
    \item Overfitting to weak label errors (across all tasks)
\end{itemize}

Our error analysis builds on the framework of debate \cite{irving2018ai}, identifying areas where adversarial dynamics can be leveraged to improve model performance and alignment.

\begin{figure}[htbp]
    \centering
    \includegraphics[width=\columnwidth]{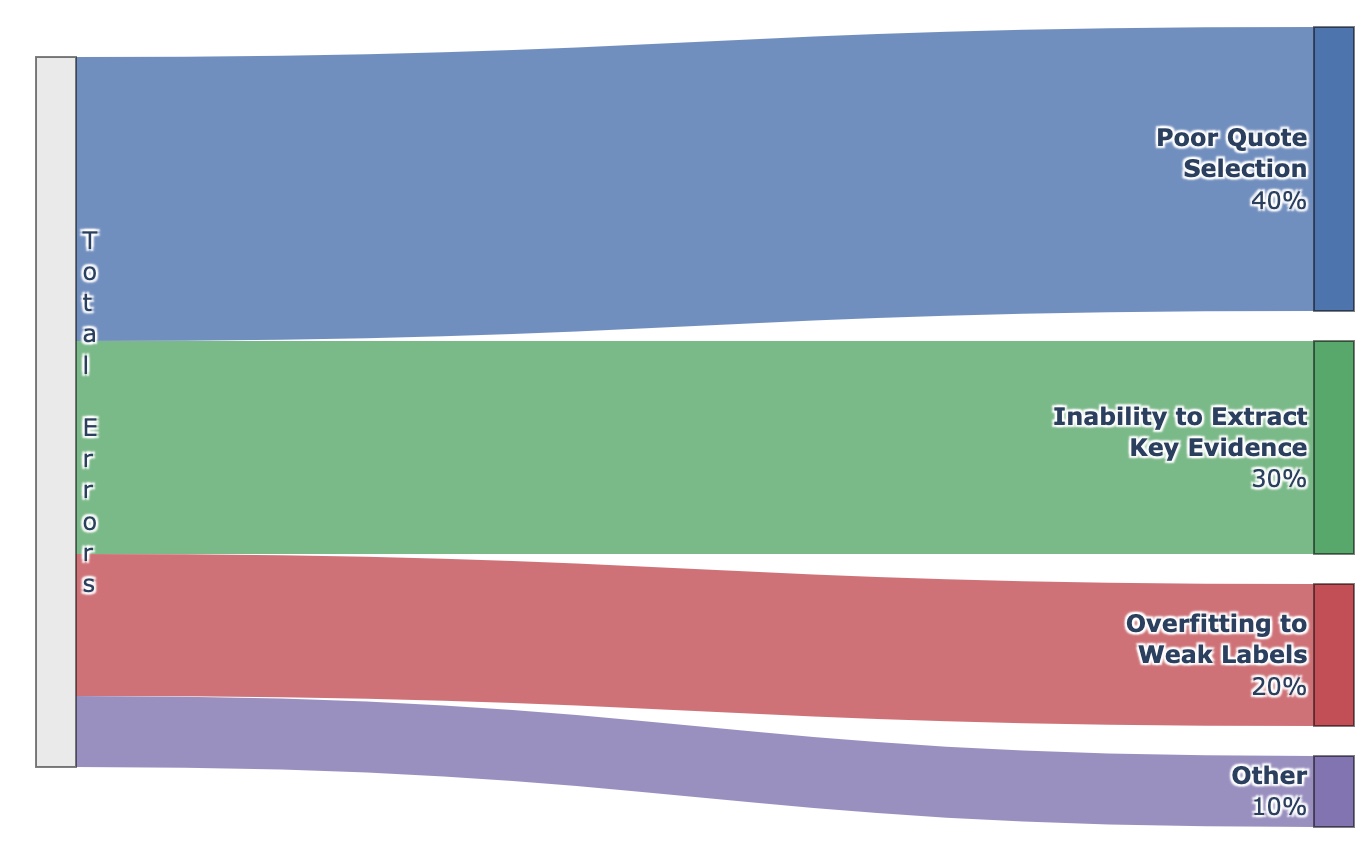}
    \caption{Distribution of error types from the error analysis. The chart shows that poor quote selection and evidence extraction account for 70\% of errors, suggesting improvements in identifying relevant information could enhance weak-to-strong learning. Overfitting errors (20\%) are relatively low, indicating the approach generally succeeds in generalizing beyond weak labels, though further improvement is possible}
    \label{fig:error-distribution}
\end{figure}

To better understand the types of errors made by the strong student model, we present the distribution of error types in Figure \ref{fig:error-distribution}.

% \subsection{Comparison to State-of-the-Art}

% We compare our best results to state-of-the-art performance on the selected NLP tasks in Table \ref{table:sota-comparison}.

% \begin{table}[htbp]
%     \centering
%     \caption{Comparison of Our Best Model's Accuracy to SOTA Performance on Key NLP Tasks. Results show our approach achieves near-SOTA performance across various linguistic challenges (CoLA: acceptability; SST-2: sentiment; MRPC: paraphrasing; QNLI: question-answering; MNLI: inference), demonstrating the effectiveness of weak-to-strong learning in competitive NLP benchmarks.}

%     \begin{tabular}{|l|c|c|}
%         \hline
%         \textbf{Task} & \textbf{Our Best} & \textbf{SOTA} \\
%         & \textbf{Accuracy} & \textbf{Accuracy} \\
%         \hline
%         CoLA & 0.85 & 0.87 \\
%         SST-2 & 0.93 & 0.95 \\
%         MRPC & 0.90 & 0.91 \\
%         QNLI & 0.89 & 0.90 \\
%         MNLI & 0.88 & 0.89 \\
%         \hline
%     \end{tabular}
%     \label{table:sota-comparison}
% \end{table}

\section{Discussion and Conclusion}

Our work demonstrates the effectiveness of weak-to-strong generalization in language models for model alignment. The proposed framework, which combines facilitation and debate-based learning, shows promise in enhancing model performance and alignment across various tasks. Through our experiments and analysis, we have gained several key insights into the nature of model alignment through facilitation:

\begin{enumerate}
    \item Strong models demonstrate a remarkable ability to generalize beyond weak supervision, particularly in NLP tasks, suggesting the potential for knowledge transfer across different capability levels.
    \item The introduction of auxiliary confidence loss significantly improves generalization, especially for large supervisor-student gaps, highlighting the importance of calibrated confidence in the alignment process.
    \item Techniques such as bootstrapping and generative finetuning show potential in addressing domain-specific challenges, as evidenced by their effectiveness in chess puzzles and reward modeling tasks.
    \item The learning process requires a delicate balance between imitation and true generalization, emphasizing the complexity of aligning AI systems with human values and intentions.
\end{enumerate}

Compared to traditional explanation generation approaches, our framework offers several advantages:

\begin{itemize}
    \item It provides a mechanism for continuous alignment as AI capabilities evolve, addressing the scalability challenges faced by current methods like RLHF.
    \item The debate-based component enhances transparency and interpretability, allowing for more nuanced oversight of the alignment process.
    \item By leveraging both weak and strong models, our approach potentially overcomes limitations of human oversight in complex tasks.
\end{itemize}

However, despite these promising results, several limitations and challenges remain, echoing concerns identified in recent alignment research \cite{ouyang2022training, bai2022constitutional, perez2020experimental}:

\begin{itemize}
    \item While an improvement over baseline approaches, naive finetuning on weak supervision is insufficient to fully recover strong model performance, indicating the need for more sophisticated transfer techniques.
    \item Generalization remains inconsistent across tasks, with complex domains like reward modeling proving particularly challenging. This highlights the need for more robust and adaptive alignment techniques.
    \item Our current setup, though an advancement over previous methods, may not fully capture the difficulties of aligning superhuman AI systems, such as the potential ease of imitating human-level errors.
\end{itemize}

These limitations point to several promising directions for future work:

\begin{itemize}
    \item Developing more sophisticated debate mechanisms to enhance the quality and efficiency of knowledge transfer between models of varying capabilities.
    \item Exploring the integration of our approach with other scalable oversight methods to create more comprehensive alignment frameworks.
    \item Investigating the application of this framework to even more advanced AI systems and diverse task domains to test its scalability and generalizability.
    \item Refining our understanding of model behavior and alignment across increasing scales of capability to better address the challenges of aligning superhuman AI systems.
\end{itemize}

In conclusion, our results demonstrate that weak-to-strong generalization is a promising approach for model alignment, capable of eliciting strong capabilities from limited supervision. By bridging the gap between explanation generation and model alignment, our framework opens new avenues for creating AI systems that are not only powerful but also fundamentally aligned with human values and intentions. While significant challenges remain in scaling this approach to more complex tasks and truly superhuman models, the insights gained from this work provide a solid foundation for future research in AI alignment and safety.

\bibliography{aaai25}

\begin{thebibliography}{26}
\providecommand{\natexlab}[1]{#1}

\bibitem[{Bai et~al.(2022)Bai, Kadavath, Kundu, Askell, Kernion, Jones, Chen, Goldie, Mirhoseini, McKinnon et~al.}]{bai2022constitutional}
Bai, Y.; Kadavath, S.; Kundu, S.; Askell, A.; Kernion, J.; Jones, A.; Chen, A.; Goldie, A.; Mirhoseini, A.; McKinnon, C.; et~al. 2022.
\newblock Constitutional ai: Harmlessness from ai feedback.
\newblock \emph{arXiv preprint arXiv:2212.08073}.

\bibitem[{Buciluǎ, Caruana, and Niculescu-Mizil(2006)}]{buciluǎ2006model}
Buciluǎ, C.; Caruana, R.; and Niculescu-Mizil, A. 2006.
\newblock Model compression.
\newblock In \emph{Proceedings of the 12th ACM SIGKDD international conference on Knowledge discovery and data mining}, 535--541.

\bibitem[{Burns et~al.(2023)Burns, Izmailov, Kirchner, Baker, Gao, Aschenbrenner, Chen, Ecoffet, Joglekar, Leike et~al.}]{burns2023weak}
Burns, C.; Izmailov, P.; Kirchner, J.~H.; Baker, B.; Gao, L.; Aschenbrenner, L.; Chen, Y.; Ecoffet, A.; Joglekar, M.; Leike, J.; et~al. 2023.
\newblock Weak-to-strong generalization: Eliciting strong capabilities with weak supervision.
\newblock \emph{arXiv preprint arXiv:2312.09390}.

\bibitem[{Cooke(2015)}]{cooke2015team}
Cooke, N.~J. 2015.
\newblock Team cognition as interaction.
\newblock \emph{Current directions in psychological science}, 24(6): 415--419.

\bibitem[{Cooke et~al.(2013)Cooke, Gorman, Myers, and Duran}]{cooke2013interactive}
Cooke, N.~J.; Gorman, J.~C.; Myers, C.~W.; and Duran, J.~L. 2013.
\newblock Interactive team cognition.
\newblock \emph{Cognitive science}, 37(2): 255--285.

\bibitem[{Datasnaek(2021)}]{lichess_chess_dataset}
Datasnaek. 2021.
\newblock Chess Game Dataset (Lichess).
\newblock Accessed: 2024-08-09.

\bibitem[{Dehghani et~al.(2017)Dehghani, Severyn, Rothe, and Kamps}]{dehghani2017learning}
Dehghani, M.; Severyn, A.; Rothe, S.; and Kamps, J. 2017.
\newblock Learning to learn from weak supervision by full supervision.
\newblock \emph{arXiv preprint arXiv:1711.11383}.

\bibitem[{Ghodratnama and Zakershahrak(2023)}]{ghodratnama2023adapting}
Ghodratnama, S.; and Zakershahrak, M. 2023.
\newblock Adapting LLMs for Efficient, Personalized Information Retrieval: Methods and Implications.
\newblock In \emph{International Conference on Service-Oriented Computing}, 17--26. Springer.

\bibitem[{Ghodratnama and Zakershahrak(2024)}]{ghodratnama2024sumrecom}
Ghodratnama, S.; and Zakershahrak, M. 2024.
\newblock SumRecom: A Personalized Summarization Approach by Learning from Users' Feedback.
\newblock \emph{arXiv preprint arXiv:2408.07294}.

\bibitem[{Gunning(2017)}]{gunning2017explainable}
Gunning, D. 2017.
\newblock Explainable artificial intelligence (xai).
\newblock \emph{Defense Advanced Research Projects Agency (DARPA), nd Web}.

\bibitem[{Hadfield-Menell et~al.(2016)Hadfield-Menell, Russell, Abbeel, and Dragan}]{hadfield2016cooperative}
Hadfield-Menell, D.; Russell, S.~J.; Abbeel, P.; and Dragan, A. 2016.
\newblock Cooperative inverse reinforcement learning.
\newblock In \emph{Advances in neural information processing systems}, 3909--3917.

\bibitem[{Irving, Christiano, and Amodei(2018)}]{irving2018ai}
Irving, G.; Christiano, P.; and Amodei, D. 2018.
\newblock AI safety via debate.
\newblock \emph{arXiv preprint arXiv:1805.00899}.

\bibitem[{LeCun, Bengio, and Hinton(2015)}]{lecun2015deep}
LeCun, Y.; Bengio, Y.; and Hinton, G. 2015.
\newblock Deep learning.
\newblock \emph{nature}, 521(7553): 436--444.

\bibitem[{Leike et~al.(2018)Leike, Krueger, Everitt, Martic, Maini, and Legg}]{leike2018scalable}
Leike, J.; Krueger, D.; Everitt, T.; Martic, M.; Maini, V.; and Legg, S. 2018.
\newblock Scalable agent alignment via reward modeling: a research direction.
\newblock \emph{arXiv preprint arXiv:1811.07871}.

\bibitem[{Miller(2018)}]{miller2018explanation}
Miller, T. 2018.
\newblock Explanation in artificial intelligence: Insights from the social sciences.
\newblock \emph{Artificial Intelligence}.

\bibitem[{Mittal et~al.(2023)Mittal, Schick, Artetxe, and Dwivedi-Yu}]{mittal2023evaluation}
Mittal, A.; Schick, T.; Artetxe, M.; and Dwivedi-Yu, J. 2023.
\newblock Evaluation of Faithfulness Using the Longest Supported Subsequence.
\newblock \emph{arXiv preprint arXiv:2308.12157}.

\bibitem[{Ng and Russell(2000)}]{ng2000algorithms}
Ng, A.~Y.; and Russell, S.~J. 2000.
\newblock Algorithms for Inverse Reinforcement Learning.
\newblock In \emph{ICML}, 663--670. Morgan Kaufmann Publishers Inc.

\bibitem[{Ouyang et~al.(2022)Ouyang, Wu, Jiang, Almeida, Wainwright, Mishkin, Zhang, Agarwal, Slama, Ray et~al.}]{ouyang2022training}
Ouyang, L.; Wu, J.; Jiang, X.; Almeida, D.; Wainwright, C.; Mishkin, P.; Zhang, C.; Agarwal, S.; Slama, K.; Ray, A.; et~al. 2022.
\newblock Training language models to follow instructions with human feedback.
\newblock \emph{Advances in neural information processing systems}, 35: 27730--27744.

\bibitem[{P{\'e}rez-D'Arpino, Khurshid, and Shah(2020)}]{perez2020experimental}
P{\'e}rez-D'Arpino, C.; Khurshid, R.~P.; and Shah, J.~A. 2020.
\newblock Experimental Assessment of Human-Robot Teaming for Multi-Step Remote Manipulation with Expert Operators.
\newblock \emph{arXiv preprint arXiv:2011.10898}.

\bibitem[{Sanh et~al.(2019)Sanh, Debut, Chaumond, and Wolf}]{sanh2019distilbert}
Sanh, V.; Debut, L.; Chaumond, J.; and Wolf, T. 2019.
\newblock DistilBERT, a distilled version of BERT: smaller, faster, cheaper and lighter.
\newblock \emph{arXiv preprint arXiv:1910.01108}.

\bibitem[{Seeber et~al.(2020)Seeber, Bittner, Briggs, De~Vreede, De~Vreede, Elkins, Maier, Merz, Oeste-Rei{\ss}, Randrup et~al.}]{seeber2020machines}
Seeber, I.; Bittner, E.; Briggs, R.~O.; De~Vreede, T.; De~Vreede, G.-J.; Elkins, A.; Maier, R.; Merz, A.~B.; Oeste-Rei{\ss}, S.; Randrup, N.; et~al. 2020.
\newblock Machines as teammates: A research agenda on AI in team collaboration.
\newblock \emph{Information \& management}, 57(2): 103174.

\bibitem[{Stiennon et~al.(2020)Stiennon, Ouyang, Wu, Ziegler, Lowe, Voss, Radford, Amodei, and Christiano}]{stiennon2020learning}
Stiennon, N.; Ouyang, L.; Wu, J.; Ziegler, D.; Lowe, R.; Voss, C.; Radford, A.; Amodei, D.; and Christiano, P.~F. 2020.
\newblock Learning to summarize with human feedback.
\newblock \emph{Advances in Neural Information Processing Systems}, 33: 3008--3021.

\bibitem[{Zakershahrak et~al.(2020{\natexlab{a}})Zakershahrak, Gong, Sadassivam, and Zhang}]{zakershahrak2020online}
Zakershahrak, M.; Gong, Z.; Sadassivam, N.; and Zhang, Y. 2020{\natexlab{a}}.
\newblock Online explanation generation for planning tasks in human-robot teaming.
\newblock In \emph{2020 IEEE/RSJ International Conference on Intelligent Robots and Systems (IROS)}, 6304--6310. IEEE.

\bibitem[{Zakershahrak et~al.(2020{\natexlab{b}})Zakershahrak, Marpally, Sharma, Gong, and Zhang}]{zakershahrak2020order}
Zakershahrak, M.; Marpally, S.~R.; Sharma, A.; Gong, Z.; and Zhang, Y. 2020{\natexlab{b}}.
\newblock Order Matters: Generating Progressive Explanations for Planning Tasks in Human-Robot Teaming.
\newblock \emph{arXiv preprint arXiv:2004.07822}.

\bibitem[{Zakershahrak et~al.(2018)Zakershahrak, Sonawane, Gong, and Zhang}]{zakershahrak2018interactive}
Zakershahrak, M.; Sonawane, A.; Gong, Z.; and Zhang, Y. 2018.
\newblock Interactive plan explicability in human-robot teaming.
\newblock In \emph{2018 27th IEEE International Symposium on Robot and Human Interactive Communication (RO-MAN)}, 1012--1017. IEEE.

\bibitem[{Ziebart et~al.(2008)Ziebart, Maas, Bagnell, and Dey}]{ziebart2008maximum}
Ziebart, B.~D.; Maas, A.; Bagnell, J.~A.; and Dey, A.~K. 2008.
\newblock Maximum Entropy Inverse Reinforcement Learning.
\newblock In \emph{Proceedings of the 23rd National Conference on Artificial Intelligence - Volume 3}, AAAI’08, 1433–1438. AAAI Press.
\newblock ISBN 9781577353683.

\end{thebibliography}

\end{document}